\pdfoutput=1
\documentclass[11pt]{article}
\usepackage[final]{acl}

\usepackage{enumitem}

\usepackage{times}
\usepackage{latexsym}
\usepackage{multirow}
\usepackage{colortbl}       % 导入colortbl包
\usepackage[T1]{fontenc}
\usepackage[utf8]{inputenc}

\usepackage{microtype}

\usepackage{inconsolata}

\usepackage{graphicx}

\usepackage{adjustbox}
\usepackage{xcolor}
\usepackage{amssymb}
\usepackage{amsfonts}
\usepackage{xspace}
\usepackage{amsmath}
\usepackage{booktabs}
\usepackage{graphicx} 
\usepackage[many]{tcolorbox}
\usepackage{hyperref}

\newcommand{\model}{CulFiT\xspace}
\newcommand{\fullmodel}{Target-aware \textbf{Cul}tural Data Synthesis and \textbf{Fi}ne-grained \textbf{T}raining\xspace}
\newcommand{\data}{GlobalCultureQA\xspace}

\newcommand{\aka}{\emph{a.k.a.,}\xspace}
\newcommand{\eg}{\emph{e.g.,}\xspace}

\newcommand{\ignore}[1]{}

\title{\model: A Fine-grained Cultural-aware LLM Training Paradigm \\ via Multilingual Critique Data Synthesis}

\author{Ruixiang Feng$^{1}$ \hspace{2mm} {\bf Shen Gao$^{1~*}$} \hspace{2mm} Xiuying Chen$^{2}$\hspace{2mm}  Lisi Chen$^{1}$\hspace{2mm} {\bf Shuo Shang$^{1}$} \thanks{Corresponding authors}\\
$^{1}$University of Electronic Science and Technology of China, \\
$^{2}$MBZUAI\\
\texttt{\texttt{\{202421081331, shengao, chenlisi\}@uestc.edu.cn}
}
\\
\texttt{xiuying.chen@mbzuai.ac.ae}, \texttt{jedi.shang@gmail.com}
}

\begin{document}
\maketitle
\begin{abstract}
Large Language Models (LLMs) have demonstrated remarkable capabilities across various tasks, yet they often exhibit a specific cultural biases, neglecting the values and linguistic diversity of low-resource regions. 
This cultural bias not only undermines universal equality, but also risks reinforcing stereotypes and perpetuating discrimination. 
To address this, we propose \model, a novel culturally-aware training paradigm that leverages multilingual data and fine-grained reward modeling to enhance cultural sensitivity and inclusivity. 
Our approach synthesizes diverse cultural-related questions, constructs critique data in culturally relevant languages, and employs fine-grained rewards to decompose cultural texts into verifiable knowledge units for interpretable evaluation. 
We also introduce \data, a multilingual open-ended question-answering dataset designed to evaluate culturally-aware responses in a global context. 
Extensive experiments on three existing benchmarks and our \data demonstrate that \model achieves state-of-the-art open-source model performance in cultural alignment and general reasoning.
\footnote{Code is available at \href{https://github.com/MMadmax/CulFiT}{https://github.com/MMadmax/CulFiT}}
% Additionally, we explore the impact of multilingual data on model robustness across languages and evaluate cultural alignment using Hofstede's cultural dimensions. 
\end{abstract}

\section{Introduction}

Large Language Models (LLMs) have demonstrated remarkable capabilities across a wide range of tasks, including reasoning~\cite{ahn2024large, huang2023large}, natural language understanding~\cite{yuan2024self, bi2024deepseek}, and daily communication. 
Owing to their advanced functionalities, LLMs have gained widespread popularity globally. 
However, these models often exhibit a Western-centric perspective~\cite{wang2023not, shen-etal-2024-understanding} and tend to neglect the values and differences of regions with low-resource languages \cite{naous-etal-2024-beer}. 
This cultural bias not only challenges the principle of universal equality but also poses significant risks, such as reinforcing stereotypes, perpetuating discrimination, and potentially inciting social conflicts. 
Consequently, there is an urgent need to develop models that are culturally sensitive and inclusive, ensuring they respect and reflect the diversity of global cultures.
To address the issue of cultural bias, recent studies~\cite{fung2024massively, nguyen2023extracting, huang2023culturally, liu2024multilingual} use LLMs to generate cultural-related texts and filter data through cleaning pipelines or human annotations. 
\citet{li2024culturepark} fine-tune culture-specific LLMs using data obtained from multi-agent communication and employ the model to tackle hate-speech detection tasks across countries.

\begin{figure}
    \centering
    \includegraphics[width=1\linewidth]{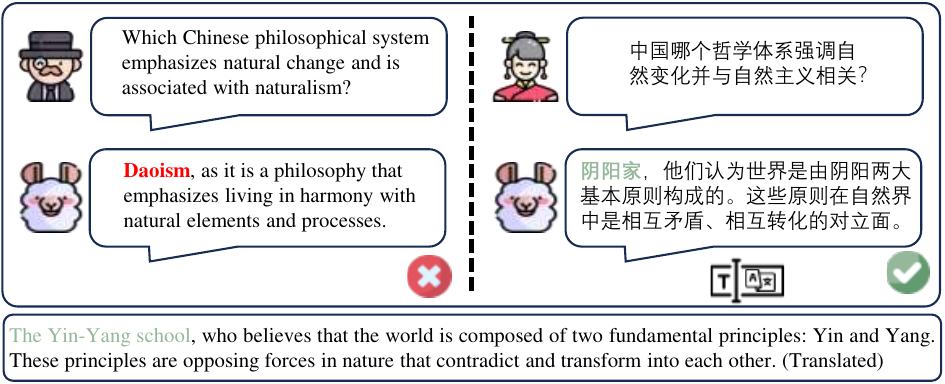}
    \caption{An example of language inconsistency. When asked cultural-specific questions, LLMs can generate correct answers in the local language but fail to provide appropriate responses in English.}
    \label{fig:intro}
\end{figure}

Previous approaches have primarily relied on descriptive, monolingual text (\eg English) to train LLMs with cultural knowledge~\cite{fung2024massively, shi2024culturebank}. 
However, understanding cultural queries often depends on the dialogue context, and such dialogue usually uses the culturally-relevant languages (\eg Malay for Singaporean culture, Chinese for Chinese culture).
Therefore, learning cultural knowledge within culturally relevant linguistic contexts is crucial. 
% Although existing models demonstrate strong performance in English-centric settings~\cite{shen-etal-2024-understanding, wang2023not}, their effectiveness significantly declines when applied to culturally-relevant languages, highlighting a lack of robustness in real-world scenarios. 
As shown in Figure~\ref{fig:intro}, when asked cultural-specific questions, LLMs can generate a correct answer in local language scenario but fail to provide appropriate responses in English, which shows a language inconsistency phenomenon. 
This phenomenon further underscores the need for culturally diverse and linguistically inclusive training approaches.
% Moreover, training on monolingual data can lead to cultural bias~\cite{li2024attributing, myung2024blend}, as the generated content tends to disproportionately reflect the cultural perspectives associated with the language of the training data. 
% This limitation further underscores the need for culturally diverse and linguistically inclusive training approaches.

Additionally, existing evaluation methods in cultural domains are coarse-grained, often relying on metrics such as text overlap, binary classification, or multiple-choice questions~\cite{chiu2024culturalbench, fung2024massively, shi2024culturebank}. 
However, these methods usually fail to account for the inherent flexibility of cultural queries~\cite{pawar2024survey}, creating a gap between the evaluation and real-world cultural knowledge applications.
These methods also lack interpretability in assessing cultural understanding, thereby undermining the reliability of the evaluation results.

In this paper, we propose \fullmodel (\model), a novel cultural-aware training paradigm that leverages target-aware multilingual data for model training and employs fine-grained rewards as training signals. 
Specifically, we first synthesize diverse cultural-related questions based on descriptive cultural knowledge texts. 
Next, we construct critique data from the content generated by the target model, which is then translated into multiple culturally relevant languages. 
This multilingual dataset is subsequently used to train the LLM. 
To provide fine-grained feedback for model training, we introduce fine-grained reward modeling, which decomposes culturally relevant texts into verifiable knowledge units, enabling a quantized interpretable evaluation of cultural alignment.

Based on the proposed data construction method, we introduce a new culturally-aware benchmark dataset \textbf{\data}, which is designed for multilingual open-ended question-answering settings, focusing on evaluating the ability to generate culturally-aware answers in a global context. 
Extensive experiments conducted on three commonly used benchmarks and \data show that our \model achieves state-of-the-art performance on open-source models.
We also explore the cultural alignment of our models based on Hofstede cultural dimensions and further investigate the effectiveness of how multi-lingual data increases robustness across various languages. 

\noindent The main contributions of this work are as follows:

\noindent $\bullet$ We propose \model, which employs multi-lingual critique data synthesis for fine-grained culturally-aware model training.

\noindent $\bullet$ We propose a target-aware data critique method to specifically address the cultural knowledge gaps in the target model, enhancing its robustness in multilingual scenarios.

\noindent $\bullet$ We introduce a fine-grained reward to quantitatively evaluate the cultural alignment.

\noindent $\bullet$ Experiments conducted on our newly proposed \data and three benchmarks demonstrate the effectiveness of \model in terms of cultural-aware metrics and general reasoning capabilities of LLM.

\section{Related Work} 

\paragraph{Cultural Bias in LLM}
% While LLMs are trained to master a vast of world knowledge \cite{achiam2023gpt}, many works have revealed that LLMs posses an unequal representation of world values across different regions and countries \cite{li2024culture, alkhamissi-etal-2024-investigating}: they often reflect a western-centric perspective \cite{wang2023not, shen-etal-2024-understanding} and overlook values from regions with low-resource languages \cite{naous-etal-2024-beer}. 
% To mitigate this, a body of research have put in an effort to enhance the cultural awareness of LLMs: through prompt based methods or tuning based methods. \citealp{choenni2024self, tao2024cultural} found that using a more culturally-aware prompt can enhance model performance, harnessing the internal culture knowledge of LLMs to improve their alignment of specific cultures.
% \citealp{li2024culturellm} leverages surveys like World Value Survey \cite{wvs} as seed questions and augment them semantically to finetune a more culturally-aware language model. 
% \citealp{li2024culturepark, feng2024modular} designed multi-agent frameworks to generate culture-related data through model communication to boost cross-culture understanding and data diversity. Deferent from existing augmenting strategies and inspired by Control Theory \cite{carver1982control}, we curate nuanced critiques to help model learn from cultural-aware answers and increase the interpretability of model outputs. 

Numerous studies have revealed that LLMs exhibit an unequal representation of world values across different regions and countries~\cite{li2024culture, alkhamissi-etal-2024-investigating}. 
Specifically, they often reflect a Western-centric perspective~\cite{wang2023not, shen-etal-2024-understanding} and overlook values from regions with low-resource languages~\cite{naous-etal-2024-beer}.
To address this issue, a growing body of research has focused on enhancing the cultural awareness of LLMs. 
For instance, \citealp{choenni2024self, tao2024cultural} found that employing culturally-aware prompts can enhance model performance by leveraging the internal cultural knowledge of LLMs. 
Similarly, \citealp{li2024culturellm} utilized surveys such as the World Value Survey~\cite{wvs} as seed questions and augmented them semantically to fine-tune a more culturally-aware model. 
% Additionally, \citealp{li2024culturepark, feng2024modular} designed multi-agent frameworks to generate culture-related data through model interactions, thereby boosting cross-cultural understanding and data diversity.
% Different from these existing augmentation strategies, and inspired by Control Theory \cite{carver1982control}, we curate nuanced critiques to help the model learn from culturally-aware answers and increase the interpretability of its outputs. 
% This approach aims to provide a more refined and contextually sensitive understanding of cultural values.

\paragraph{Cultural Data Synthesis} 
% There has been substantial progress in the development of datasets related to cultural aspects. \citealp{huang2023culturally, lee2024exploring}construct their cultural dataset through human annotation, which is labor-intensive and hard to scale. Meanwhile, many work develop data cleaning pipelines and filter data from social media such as Tiktok and Reddit \cite{shi2024culturebank, nguyen2023extracting} or Wikimedia \cite{fung2024massively, liu2024multilingual} to cover a wide range of cultural knowledge. \citealp{rao2024normad, shum2023automatic} synthesize their data from existing datasets, and transfer the cultural knowledge to specific area such as norms or etiquette. Nevertheless, most cultural datasets are predominantly composed in English and thus may not effectively capture the contextual nuances of real-world settings.

Significant progress has been made in the development of datasets related to cultural aspects. 
\citealp{huang2023culturally, lee2024exploring} construct the cultural datasets through human annotation, which is labor-intensive and difficult to scale. 
Meanwhile, many works develop data cleaning pipelines from social media platforms such as TikTok, Reddit~\cite{shi2024culturebank, nguyen2023extracting} and Wikimedia~\cite{fung2024massively, liu2024multilingual}. % to cover a broad range of cultural knowledge. 
\citealp{rao2024normad, shum2023automatic} synthesize their data from existing datasets and transfer cultural knowledge to specific domains such as norms or etiquette. 
However, most cultural datasets are predominantly composed in English, limiting their ability to effectively capture the context of real-world scenarios.

\paragraph{Cultural Benchmarks} 
% To investigate and ease the cultural bias that exists in LLM, extensive research has been dedicated to building cultural benchmarks, which can be categorized into two types: cultural specific benchmarks and multicultural benchmarks. First, culture-specific benchmarks are developed to evaluate LLM's cultural capacities in specific geographical regions and countries, including South East Asia \cite{wang2023seaeval}, China \cite{sun2024benchmarking}, and  Arabic regions \cite{mousi2024aradice}, etc. 
% On the other hand, multicultural benchmarks explore cultural diversity from a broader perspective, whether through human annotating~\cite{chiu2024culturalbench, myung2024blend}, model generating~\cite{putri2024can} and AI-human assistance~\cite{chiu2024culturalteaming}. 
% However, previous benchmarks are mainly in the form of multiple-choice or Yes/No questions, which suffer from positional bias and lack fine-grained evaluation in open-ended scenarios.

Extensive research has also focused on developing cultural benchmarks, which can be categorized into: culture-specific benchmarks and multicultural benchmarks. 
Culture-specific benchmarks are designed to evaluate LLMs' cultural capacities in specific regions and countries, such as Southeast Asia~\cite{wang2023seaeval} and China~\cite{sun2024benchmarking}. %, and Arabic regions~\cite{mousi2024aradice}. 
On the other hand, multicultural benchmarks aim to explore cultural diversity, constructed by human annotation~\cite{chiu2024culturalbench, myung2024blend}, model generation~\cite{putri2024can}, and human in the loop~\cite{chiu2024culturalteaming}. 
However, these methods primarily rely on multiple-choice or Yes/No questions, which are prone to positional bias and lack fine-grained evaluation in open-ended scenarios.

% To address this, we propose (XXX), a new multicultural benchmark consisting of 1104 open-ended QA pairs that evaluates LLM's cultural knowledge answering ability in multilingual settings. Despite this, we also design a novel evaluation metric that separates each answer into small but verifiable units, 
% evaluating the cultural awareness in detail.

% \paragraph{Cultural datasets and benchmarks}
% Significant progress has been made in the development of datasets and benchmarks related to cultural aspects. 
\section{\model}
% In this section, we detail the methodology of our proposed \fullmodel (\model).

\subsection{Overview}

% Cultural knowledge is implicitly hidden in real-life scenarios and differs in languages, making it hard to capture it in nuance and scale in local settings. 
% To address the gap between real-world cultural usage  and existing cultural knowledge implementations, we propose XXX by multilingual cultural data augmentation with minimal human intervention. 
% As shown in Figure X, we first show how to collect our QA data from multiple datasets \S\ref{sec:QA Synthesis}, followed by nuanced critique generation \S\ref{sec:Critique Generation}, data translation process \S\ref{sec: Translation} and finally two-stage training method\S\ref{sec: Training}.

\begin{figure*}
    \centering
    \includegraphics[width=1\linewidth]{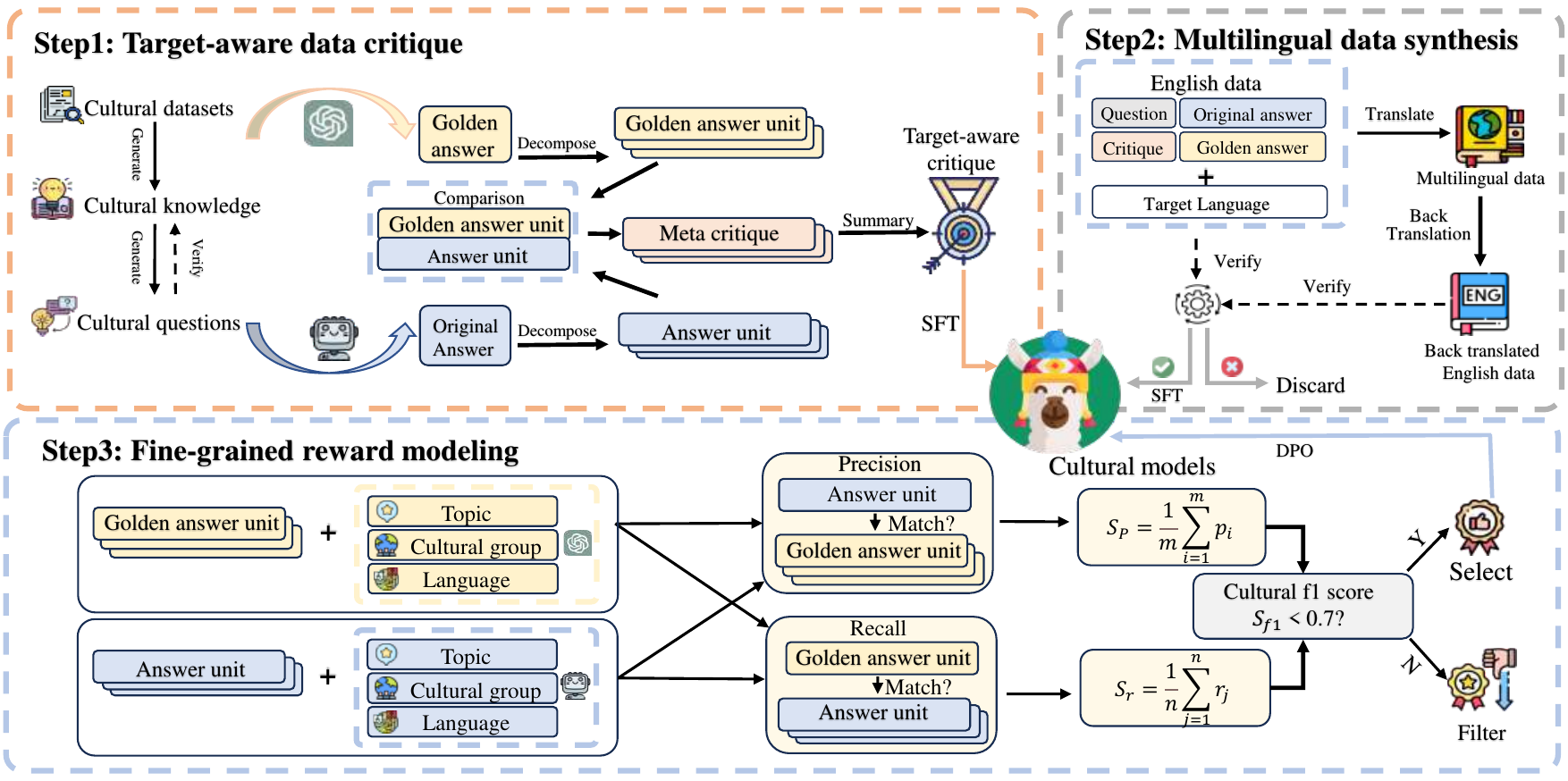}
    \caption{The overview of our proposed \model.}
    \label{fig:main}
\end{figure*}

% Existing approaches~\cite{} typically train cultural-aware models using descriptive texts while neglecting cross-lingual contextualization in real-world applications, resulting in fragile generalization across linguistic environments. 
% In this paper, we propose \fullmodel (\model), a framework that trains the cultural-aware LLM using multilingual cultural data to mitigate cultural biases in language models where the multilingual data is synthesized with target-aware critique.
As illustrated in Figure~\ref{fig:main}, our \fullmodel (\model) comprises three components: 
(1) \textbf{Target-aware critique generation} reflects the common errors for the target LLM (\S\ref{sec:target-critique}).
(2) \textbf{Multilingual data synthesis} increases the generalization ability in real-world application scenarios by augmenting the cultural-aware data (\S\ref{sec:multi-lingual-synthesis}).
(3) \textbf{Fine-grained model training} provides interpretable evaluation protocols to optimize the LLM (\S\ref{sec:training}).

\subsection{Target-aware Data Critique} \label{sec:target-critique}
% To construct geo-diverse and culturally-aware QA data, we use CANDLE, CultureAtlas and Culturebank as our source data for their quality and open-source nature. 
% However, the cultural texts in above datasets are discrete and only consist of assertive statements, which hinder the nature that cultural points often appear in routine questions. To mitigate this, we first combine the discrete cultural texts from same topics and use gpt-4o-mini to convert them into a paragraph of fluent, coherent cultural knowledge, denoted as $K$. After that, we prompt LLM to generate culture-relevant questions based on the cultural knowledge $K$ and verify the containment of $K$ and the quality of cultural questions, denoted as $Q$. For the answer part, we collect golden answers $A_g$ by integrating culture knowledge $K$ and leveraging stronger models like gpt-4o, which gives the theoretical upper-bound cultural-awareness performance of our questions. Meanwhile, we also use weaker models like Llama3.1-8B-Instruct and Qwen2.5-7B-Instruct to generate lower quality answer $A_t$ that will be further criticized. During this process, we use human-written few-shots prompting method to ensure instruction following and answer quality.

\paragraph{Data Synthesis}

We first construct cultural-aware QA pairs from three widely-used data sources: CANDLE~\cite{nguyen2023extracting}, CultureAtlas~\cite{fung2024massively}, and CultureBank~\cite{shi2024culturebank}.
However, these datasets primarily contain discrete, assertive statements that fail to reflect how cultural concepts naturally emerge when chatting with users. 
To address this limitation, we first aggregate related cultural statements by topic and synthesize them into coherent knowledge paragraphs $K$ by employing a data generation model $\mathcal{G}$. 
We then employ prompting strategies to generate culturally-grounded questions $Q$ based on the knowledge $K$, with automated verification by using $\mathcal{G}$ to ensure each question is answerable using $K$.
Then we generate two type of answers with two different LLMs: 
(1) \textit{Golden Answer} ($A_g$): Produced by data generation LLM $\mathcal{G}$ through knowledge-aware synthesis.
(2) \textit{Target-aware Answers} ($A_t$): Generated by the target model $\mathcal{M}$ using few-shot exemplars to control answer quality and instruction following, where the target model $\mathcal{M}$ denotes the model which we want to fine-tune.

\paragraph{Critique Generation} \label{sec:Critique Generation}

% Inspired by Control Theory, which posits that self-regulation and discrepancy-reducing feedback is beneficial for building social personality and cultural psychology, we design nuanced critique generation pipeline that leads smaller model to reach the upper-bound answer $A_g$ from original answer $A_t$. Nevertheless, recent research revealed that existing LLMs still struggle to directly provide insightful critique\cite{huang2023large, kamoi2024can} and tend to turn right responses into wrong ones\cite{gou2023critic}, thus it is crucial to ensure the quality of the proposed critique\cite{sun-etal-2024-critique}.To achieve this, we break down the golden answer $A_g$ and the answer-to-critique $A_t$ into small knowledge points that each contains a piece of information mentioned by the corresponding answer. Therefore, we can get two lists of information units that represents the golden answer and answer-to-critique, denoted as $A_g^u=[A_g^1, A_g^2, \cdots , A_g^n]$ and $A_t^u=[A_t^1,A_t^2, \cdots, A_t^m]$, where $n$ and $m$ indicates the length of the representing units.

Inspired by control theory in sociology~\cite{carver1982control}, which posits that self-regulation and discrepancy-reducing feedback contribute to the development of social identity and cultural cognition, we propose a critique-based data generation framework for targeted cultural knowledge acquisition. 
However, recent studies~\cite{huang2023large,kamoi2024can} reveal that conventional critique generation methods often fail to provide insightful feedback for improving cultural knowledge. 
Moreover, \citet{gou2023critic} demonstrates that simply using direct-generated critique can degrade model performance by corrupting correct responses. 
% These findings underscore it is crucial to ensure the quality of critique~\cite{sun-etal-2024-critique}.
To address these challenges, we propose to decompose the golden answer $A_g$ and target-aware answers $A_t$ into atomic cultural knowledge units. 
This decomposition yields two knowledge units sequences: $A_g^u = [A_g^1, A_g^2, \cdots, A_g^n]$ and $A_t^u = [A_t^1, A_t^2, \cdots, A_t^m]$, where $n$ and $m$ denote the sequence lengths representing distinct cultural knowledge units:
\begin{align}\label{equ:knowledge-decompose}
    A_g^u &= \mathcal{G}(A_g)=[A_g^1, A_g^2, \cdots, A_g^n]  ,\\
    A_t^u &= \mathcal{G}(A_t)=[A_t^1, A_t^2, \cdots, A_t^m],
\end{align}
where $\mathcal{G}$ denotes the generation model same above.

% After having the knowledge point units that substitute the direct answers, we can obtain nuanced critique tuples $T$ by comparing each golden answer unit ${A_g^i} \in A_g^u$ with answer-to-critique unit $A_t^j \in A_t^u$.  Notably, each tuple $T_i \in T$ consists of three items and can be presented as: \begin{equation}
%     T_i = \{A_g^i, A_t^j, C_r\}
% \end{equation}where $C_r$ denotes the meta critique. We also let meta critique falls into three types that can further control the quality of our critiques:\begin{enumerate}[label=(\arabic*), left=2em]
%     \item Roughly the same: this means the golden answer unit $A_g^i$ we are processing finds a same meaning unit in original answer points $A_t^j \in A_t^u$, and thus we don't need to conduct refinement on this cultural knowledge point.
%     \item Not addressed clearly: this shows that current golden answer unit $A_g^i$ cannot be found in any answer point $A_t^j \in A_t^u$, so we need to clearly point out the missing cultural point in our meta critique.
%     \item Contradicted: this suggests that present golden answer unit $A_g^i$ finds a contradictory statement in original answer points $A_t^j \in A_t^u$, which should be stated in the corresponding meta critique $C_r$ that this answer knowledge point needs to be corrected to adhere the real cultural customs and norms.
% \end{enumerate}

After obtaining the knowledge units $A_g^u$ and $A_t^u$, we construct a fine-grained critique set $T$ by comparing each ground truth knowledge unit ${A_g^i} \in A_g^u$ with corresponding knowledge units $A_t^j \in A_t^u$ in the target-aware answer. 
To ensure critique quality, we generate the meta-critique $C_r$ by the data generation model $\mathcal{G}$, and categorize meta-critique $C_r$ into three types:

(1) \textit{Semantic Equivalence}: Indicates $A_g^i$ has an exact semantic match in $A_t^u$, suggesting no further training is required for this cultural knowledge unit.

(2) \textit{Unaddressed Knowledge}: Occurs when $A_g^i$ lacks any corresponding unit in $A_t^u$, necessitating explicit pointing out this cultural knowledge in $C_r$.

(3) \textit{Contradictory Statement}: Identifies cases where $A_g^i$ conflicts with statements in $A_t^u$, requiring corrective meta-critique $C_r$ to align the target model $M$ with appropriate cultural norms.

This critique method allows the model to compare the golden answer $A_g$ with target-aware answers $A_t$, thereby generating nuanced and reliable targeted critiques. These critiques will direct subsequent supervised fine-tuning to prioritize knowledge domains where the target LLM is prone to errors.
Each critique instance $T_i \in T$ is represented as a triple:
\begin{equation}
T_i = \{A_g^i, A_t^j, C_r\},
\end{equation}
where $C_r$ denotes the meta-critique described above. 
Finally, we summarize all meta critiques $(T_1, T_2, \cdots, T_k)$ for corresponding answer into a comprehensive critique $C$, and it will be used to serve as target-aware cultural error reminder in the supervised fine-tuning stage.
\begin{equation}
    C = \text{LLM}(P, (T_1, T_2, \cdots, T_k)),
\end{equation}
where $P$ denotes the critique summary prompt and $C$ denotes the final critique we obtain. All prompts can be seen in Appendix \S~\ref{subsec: prompts}

% Overall, the critique points generation process which can be described as:
% \begin{equation}
%     (T_1, T_2, \cdots, T_k) = Model(P, A_g^u, A_t^u)
% \end{equation}
% where $P$ is the prompt we deliver into language model and $k$ is the length of the critique tuples and theoretically should be the same as the length of golden answer units.

\subsection{Multi-lingual Data Synthesis} \label{sec:multi-lingual-synthesis}

% After collecting cultural QA pairs with critique, we first translate them into local languages using gpt-4o-mini, and then use back-translation technique to verify the quality of the translation, which involves in prompting LLM to compare the translated text with the original texts and filtering the lower-quality translation text. 
% The motivations behind our translation method are primarily two folds: 
% (1) Existing LLMs even the SOTA LLMs still lack the capacity to directly answer cultural questions in multilingual settings, especially in low-resource settings, so we take a simpler translation task instead of generation task. 
% (2) Augmenting data in multilingualism can imitate how local residence communicate, thus results in a more close real-world cultural usage and increases the diversity and robustness \S\ref{sec:Experimental Results}.

In real-world scenarios, cultural-aware dialogue frequently occur in culturally-relevant languages (\eg Malay for Singaporean culture, Chinese for Chinese culture). 
To enhance model robustness in generating culturally appropriate knowledge across multilingual contexts, we propose a \textit{Multi-lingual Data Synthesis} approach that generates answers in culturally relevant languages. 
After collecting critique-annotated cultural data $U = (Q, A_g, A_t, C)$, we first translate the data into target languages using our data generation model $\mathcal{G}$:
\begin{equation}
U_{target} = \mathcal{G} (U, L),
\end{equation}
where $U$ and $U_{target}$ represent the source and target language cultural data, respectively, and $L$ denotes the target language.
To mitigate hallucination and ensure translation quality, we employ a back-translation verification mechanism. 
Specifically, we translate the target language text back to English using generation model $\mathcal{G}$:
\begin{equation}
U_{back} = \mathcal{G}(U_{target} \rightarrow U),
\end{equation}
where $U_{back}$ represents the back-translated text. 
We then perform semantic alignment between $U_{back}$ and the original English text $U$, ensuring 
% only multilingual data pairs that demonstrate consistent semantic meaning.
consistent semantic meaning of multilingual pairs.
 
\subsection{Fine-grained Model Training} \label{sec:training}

To train a cultural-aware model, we conduct a two-stage training method, which first uses supervised fine-tuning with target-aware multi-lingual critique data to equip the model with the ability to rectify areas prone to errors in the original answer and then leverage Direct Preference Optimization (DPO)~\cite{rafailov2024direct} to further align the model. 
However, due to the challenge of intricacy and reliability in rewarding cultural-related texts, in this paper, we propose a \textit{fine-grained cultural-aware reward modeling} approach that contains two sub-metrics: cultural precision and cultural recall to evaluate how culturally reliable the open-ended answer is.

Firstly, we enhance the evaluation framework by requiring the model to generate three additional contextual units for each question: 
(1) cultural group affiliation ($A^c$), 
(2) cultural topic ($A^s$), and 
(3) primary language(s) of the cultural group ($A^l$). 
These units are appended to the original answer units, forming an extended answer representation:
\begin{equation}
A = [A^1, A^2, \cdots, A^k, A^c, A^s, A^l],
\end{equation}
where $A^{1-k}$ denotes the atomic answer units obtained in \ref{sec:target-critique}.
The inclusion of these contextual units serves two purposes. 
First, it captures the cultural contextual awareness, since as a culturally-aware model should accurately identify the background of the question. 
Second, these contextual units $A^c, A^s, A^l$ provide precise, easily verifiable evaluation targets that reduce scoring variance due to their concise and factual nature.

% We then use the reward function described in \ref{sec: Evaluation Metric} that calculated a cultural f1 score $S_{f1}$ based on the fine-grained reward modeling method. 

% Based on empirical analysis of training performance, we select preference data with $S_{f_1} \leq 0.7$, resulting in a curated training set of 16,779 preference pairs in the DPO stage.

% 你这节叫modeling training，上面构造完A之后就没了，这个A干啥用，model咋training都没说，这部分讲的时候应该说我们用了一个reward，然后引出下面那一个section

\subsubsection{Fine-grained Reward Modeling} \label{sec: Evaluation Metric}

% 使用我们生成的数据来FT LLM，那些是输入，那些是groundtruth。为了xxx，我们使用DPO，为了xx，我们使用一个xx的reward。
% We use above synthesized data to fine-tune our LLM, 
During our training process, we begin by applying supervised fine-tuning that takes questions $Q$ combined with original answer $A_t$ and target-aware critique $C$ as input and grounded answer $A_g$ as output. To align the model with  human cultural preferences, we also adopt Direct Preference Optimization. However, human preference is often subjective and context-dependent, which is hard to quantify in a reward function.  To address the gap between the inherently subjective nature of cultural judgments and the objective metrics provided by standard reward functions, we design a fine-grained reward function that fully assesses the quality of cultural answers to select preference pairs automatically and robustly.
% Existing cultural research fail to quantify the open-ended cultural questions due to its intricacy and difficulty to capture cultural nuances. 
% Inspired by \cite{sun-etal-2024-critique}, we define cultural precision task and cultural recall task to reliably evaluate how cultural the open-ended answer is. 

\paragraph{Cultural Precision Metric}

% We design cultural precision task to evaluate the incorporation scope of the cultural knowledge in the proposed answer. 
% Intuitively, a more culturally-aware answer should precisely contain the relevant cultural knowledge, thus hitting the correct cultural knowledge in the golden answer can be seen as a positive signal when evaluating the answer. 
% Similar to \S\ref{sec:Critique Generation}, we first divide the golden answer and proposed answer into small but verifiable units, known as $A_g^u=[A_g^1, A_g^2, \cdots , A_g^n]$ and $A_t^u=[A_t^1,A_t^2, \cdots, A_t^m]$, and then use LLMs to judge whether proposed answer knowledge units correlate with the golden answer knowledge units. 
% Specifically, the precision calculation process of proposed answer unit $A_t^i \in A_t^u$ can be described as: 
% \begin{equation}
% p_i =
%  \begin{cases}1 & \text{if } A_t^i\text{ matches any } A_g^j \in A_g^u,\\
%  0 & \text{otherwise }
%  \end{cases} 
% \end{equation}
 
% After this, We then can calculate the overall precision score:\[S_p = \frac{\sum\limits_{i=1}^m p_i}{m}\]

We introduce a \textit{Cultural Precision Metric} $\mathbf{M_p}$ to evaluate the extent of cultural knowledge incorporation in model-generated answers. 
The intuition for this metric is that culturally-aware responses should precisely encompass relevant cultural knowledge. 
Following the methodology in Equation~\ref{equ:knowledge-decompose}, we decompose both golden and target-aware answers into verifiable knowledge units, denoted as $A_g^u = [A_g^1, A_g^2, \cdots, A_g^n]$ and $A_t^u = [A_t^1, A_t^2, \cdots, A_t^m]$, respectively. 
The precision evaluation for each proposed answer unit $A_t^i \in A_t^u$ is formalized as:
\begin{equation}
p_i =
 \begin{cases}
   1 & \text{if } \exists A_g^j \in A_g^u \text{ where } A_t^i \text{ matches } A_g^j,\\
   0 & \text{otherwise}
 \end{cases} 
\end{equation}
The cultural precision $S_p$ is then computed as:
\begin{equation}
\begin{aligned}
S_p &= \mathbf{M_p}(P) = \frac{1}{m}\sum\limits_{i=1}^m p_i
\end{aligned}
\end{equation}

\paragraph{Cultural Recall Metric} 

% We also design cultural recall task to assess the coverage of the golden answer over the proposed answer. The motivation behind this task is that if the proposed answer contains all the golden answer points, at least this answer can be seen as a "culture-completed" answer, which means it is broad enough to cover all the culture nuances of the question. Similar to precision task, we can calculate each unit's recall score as below:
% \begin{equation}
% r_i =
%  \begin{cases}1 & \text{if } A_g^j\text{ matches any } A_t^i \in A_t^u,\\
%  0 & \text{otherwise }
%  \end{cases} 
% \end{equation}
% and then the overall recall score:
% \begin{equation}
%  S_r = \frac{\sum\limits_{j=1}^n r_j}{n}
% \end{equation}

To complement the precision evaluation, we introduce a \textit{Cultural Recall Metric} $M_r$ that measures the coverage of golden answer knowledge units in the proposed answer. 
This metric is motivated by the principle that a comprehensive cultural response should encompass all relevant cultural knowledge points present in the golden answer, thereby achieving ``culture-completeness''. 
Following the same unit decomposition approach as in the precision task, we evaluate recall at the knowledge unit level through pairwise matching.
The recall score for each golden answer unit $A_g^j \in A_g^u$ is computed as:
\begin{equation}
r_j =
 \begin{cases}
   1 & \text{if } \exists A_t^i \in A_t^u \text{ where } A_g^j \text{ matches } A_t^i,\\
   0 & \text{otherwise}
 \end{cases} 
\end{equation}
The cultural recall score $S_r$ is then calculated as:
\begin{equation}
\begin{aligned}
S_r &= \mathbf{M_r}(R) = \frac{1}{n}\sum\limits_{j=1}^n r_j
\end{aligned}
\end{equation}

\paragraph{Cultural F1 Metric} 
 
% To integrate precision task and recall task into a quantitative score, we introduce culture F1 score, which is the harmonic mean of the precision score $S_p$ and recall score $S_r$:
%  \begin{equation}
%      S_{f_1} = 2 \cdot \frac{S_p \cdot S_r}{S_p + S_r}
%  \end{equation}
% During the implementation of this metric, we also let model generate three more answer units related to the question: 
% (1) the cultural group of regarding to the question, 
% (2) the topic of the question, 
% (3) the language(s) mainly speak in the cultural group. 
% We then append them to the last of the answer unit list, denoted as $A = [A^1, A^2, \cdots, A^n, A^c, A^s, A^l]$, where $A^{1-n}$ is the answer units derived from the answer before and $A^c$, $A^s$, $A^l$ represents the model-generated cultural group, cultural topic and language, respectively. 
% We add these additional units is because that a more cultural model should discern the background of the question, thinking in a correct background should also be seen as a positive reward to our model. 
% Despite this, these three units are relatively short and can be evaluated more precisely, thus reduce the disturbance in the evaluation process. 
% We choose the preference data with $Cultural F1 Score \leq 0.7$, for the reason of training performance and thus construct 16779 preference pairs for training.

To provide a comprehensive evaluation metric, we introduce the \textit{Culture F1 Metric}, which combines precision and recall through their harmonic mean:
\begin{equation}
S_{f_1} = 2 \cdot \frac{S_p \cdot S_r}{S_p + S_r}.
\label{Eq: threshold}
\end{equation}
It is notable that we select our DPO training data using cultural F1 metric with $S_{f1}$ < 0.7, which results in preference pairs maintaining both high-quality and relative larger learnable cultural gaps.

\section{Experimental Setup}

\subsection{Datasets} 

We test our \model and baselines on four datasets.
\data is our newly proposed multilingual benchmark that evaluates open-ended cultural knowledge question answering ability with 1104 questions, covering 400 specific topics and 23 languages. 
CANDLE500~\cite{nguyen2023extracting} and CulturalBench~\cite{chiu2024culturalbench} are multi-choice benchmarks that focus on evaluating cultural knowledge with 500 and 1224 samples. 
BLEnD~\cite{myung2024blend} is a hand-crafted benchmark designed to evaluate LLM's cultural common knowledge across 16 countries and 13 different languages, comprising 52.6k question-answer pairs. 
Detailed distribution of topics and cultural groups across continents of \data and examples are in Appendix \S~\ref{subsec: GlobalCultureQA}, \S~\ref{subsec: Case Study}. 
% LLMs are required to answer cultural questions in the formats of short answers in this dataset, with prompts in local languages.  

\subsection{Baselines} 

We employ several state-of-the-art LLM as baselines: close-source models including gpt-4o (\texttt{4o}) and gpt-4o-mini (\texttt{4o-mini})~\cite{hurst2024gpt} and open-source models including Llama3.1-8B-Instruct (\texttt{Llama3.1})~\cite{dubey2024llama}, Qwen2.5-7B-Instruct (\texttt{Qwen2.5})~\cite{yang2024qwen2}, mistral-7B-Instruct-v0.3 (\texttt{Mistral}), aya-8B-expanse (\texttt{Aya}), SeaLLMs-v3-7B-Chat (\texttt{SeaLLMs}) and \texttt{CultureBank}~\cite{shi2024culturebank}.
Training details can be found in \S~\ref{subsec: training detail}.
% ~\footnote{\url{https://huggingface.co/mistralai/Mistral-7B-Instruct-v0.3}}
% ~\footnote{\url{https://huggingface.co/CohereForAI/aya-expanse-8b}}
% ~\footnote{\url{https://huggingface.co/SeaLLMs/SeaLLMs-v3-7B-Chat}}

\subsection{Evaluation Metric} 

For \data benchmark, we evaluate cultural precision score $S_p$, cultural recall score $S_r$ and then calculate cultural f1 score $S_{f1}$ described in \S~\ref{sec: Evaluation Metric}. 
For CANDLE500 and CulturalBench, we report the precision of multi-choice questions. 
As for BLEnD, we first use corresponding lemmatizers and stemmers for model-generated answers and then compute the scores by marking whether the LLM's answer is included by the human annotator's answer. 

\section{Experimental Results} \label{sec:Experimental Results}
\begin{table}
    \centering
    \resizebox{0.9\columnwidth}{!}{
    \begin{tabular}{l|c|c|c}
    \toprule
    \textbf{Model} & \textbf{Precision} & \textbf{Recall} & \textbf{F1} \\
    \midrule
    \multicolumn{4}{c}{Close-source Models} \\
    \midrule
    \texttt{4o} & 72.34 & \textbf{73.29} & 72.81 \\
    \texttt{4o-mini} & 72.89 & 72.47 & 72.68 \\
    \midrule
    \multicolumn{4}{c}{Open-source Models} \\
    \midrule
    \texttt{Mistral} & 67.26 & 68.76 & 66.73 \\
    \texttt{SeaLLMs} & 71.50 & 66.04 & 68.71 \\
    \texttt{Aya} & 69.88 & 69.52 & 68.66 \\
    \midrule
    \texttt{Qwen2.5} & 66.97 & 68.80 & 66.79 \\
    \model (Qwen2.5) & 72.16 & 67.56 & 68.81 \\
    \textit{-SFT} & 69.11 & 67.44 & 68.26\\
    \textit{-DPO} & 70.95 & 66.57 & 67.60 \\
    \midrule
    \texttt{Llama3.1} & 62.52 & 68.96 & 64.53 \\
    \model (Llama3.1) & \textbf{74.73} & 71.21 & \textbf{72.94} \\
    \textit{-SFT} & 71.33 & 70.55 & 70.94\\
    \textit{-DPO} & 74.07 & 69.84 & 70.81 \\
    \bottomrule
    \end{tabular}
    }
    \caption{Performance on our \data.}
    \label{tab:openculture}
\end{table}
\begin{table}
\centering
% \small
\resizebox{0.9\columnwidth}{!}{
\begin{tabular}{l|c|c}
\toprule
\textbf{Model} & \textbf{CANDLE500} & \textbf{CulturalBench} \\
\midrule
\multicolumn{3}{c}{Close-source Models} \\
\midrule
\texttt{4o} & 91.2 & 84.1 \\
\texttt{4o-mini} & 87.0 & 82.1 \\
\midrule
\multicolumn{3}{c}{Open-source Models} \\
\midrule
\texttt{Mistral} & 69.0 & 67.1 \\
\texttt{Aya} & 73.2 & 67.2 \\
\texttt{SeaLLMs} & 75.2 & 68.5 \\
\texttt{CultureBank} & 38.4 & 53.8 \\
\midrule
\texttt{Qwen2.5} & 76.0 & 68.9 \\
\model (Qwen2.5) & 79.6 & 72.9 \\
    \textit{-SFT} & 76.4 & 70.9\\
    \textit{-DPO} & 78.2 & 71.0  \\
\midrule
\texttt{Llama3.1} & 72.4 & 66.5 \\
\model (Llama3.1) & \textbf{81.2} & \textbf{73.1} \\
    \textit{-SFT} & 78.6 & 69.1\\
    \textit{-DPO} & 80.0 & 71.9 \\
\bottomrule
\end{tabular}
}
\caption{Performance on CANDLE500 and CulturalBench.}
\label{multi-choice}
\end{table}

\begin{table*}[t]
\centering
\resizebox{1\textwidth}{!}{
\begin{tabular}{l|cccccccccccccccc}
\toprule 
 \textbf{Models} & \textbf{US} & \textbf{GB} & \textbf{CN} & \textbf{ES} & \textbf{MX} & \textbf{DZ} & \textbf{GR} & \textbf{KR}  & \textbf{JB} & \textbf{IR} & \textbf{ID} & \textbf{AZ} & \textbf{KP} & \textbf{NG} & \textbf{AS} & \textbf{ET} \\ 
 \midrule

 \multicolumn{17}{c}{Close-source Models} \\
 \midrule
\texttt{4o} & 84.29 & 82.37 & 76.48 & 78.46 & 76.36 & 60.36 & 65.34 & 66.36 & 55.83 & 70.56 & 69.46 & 59.48 & 45.98 & 40.26 & 43.67 & 20.51 \\
\texttt{4o-mini} & 83.72 & 82.78 & 73.51 & 77.34 & 76.48 & 59.34 & 66.87 & 53.61 & 69.72 & 69.12 & 68.13 & 49.57 & 43.39 & 39.48 & 40.69 & 17.25 \\
\midrule
\multicolumn{17}{c}{Open-source Models} \\
 \midrule
\texttt{Mistral} & 83.29 & 82.41 & 48.31 & 60.12 & 58.24 & 30.20 & 25.09 & 48.0 & 8.21 & 33.77 & 60.83 & 27.93 & 35.58 & 12.47 & 8.80 & 3.95 \\
\texttt{SeaLLMs} & 77.09 & 73.01 & \textbf{66.97} & 64.39 & 64.30 & 37.98 & 21.37 & 45.89 & 20.85 & 30.04 & 51.87 & 24.09 & 34.23 & 8.35  & 10.69 & 3.17 \\
\texttt{Aya}   & 81.56 & 76.26 & 54.36 & 62.78 & 57.75 & \textbf{47.95} & \textbf{47.33} & 54.32 & 19.24 & 46.83 & \textbf{66.64} & 24.52 & 34.23 & 16.72 & 12.99 & 10.77 \\
\midrule
\texttt{Llama3.1} & 82.46 & 76.48 & 56.54 & 61.62 & 64.09 & 40.73 & 41.52 & 50.94 & 12.43 & 48.46 & 58.75 & 39.87 & 36.26 & 20.42 & 15.09 & 8.26 \\
\model (Llama3.1) & \cellcolor{green!20}\textbf{85.46} & \cellcolor{green!20}\textbf{83.29} & \cellcolor{green!20}57.38 & \cellcolor{green!20}\textbf{67.59} & \cellcolor{green!20}\textbf{66.17} & \cellcolor{green!20}45.76 & \cellcolor{green!20}42.17 & \cellcolor{red!20}49.16 & \cellcolor{green!20}20.13 & \cellcolor{green!20}\textbf{49.78} & \cellcolor{green!20}64.87 & \cellcolor{green!20}\textbf{43.92} & \cellcolor{green!20}\textbf{37.61} & \cellcolor{green!20}21.03 & \cellcolor{green!20}\textbf{21.80} & \cellcolor{green!20}\textbf{12.34} \\
\midrule
\texttt{Qwen2.5} & 78.52 & 72.52 & 60.33 & 64.60 & 58.24 & 38.15 & 21.05 & 52.19 & 21.17 & 36.18 & 49.58 & 38.59 & 25.67 & 24.60  & 19.25 & 11.41 \\
\model (Qwen2.5) & \cellcolor{green!20} 80.12 & \cellcolor{green!20} 77.48 & \cellcolor{green!20}63.28 & \cellcolor{green!20}66.78 & \cellcolor{green!20}60.36 & \cellcolor{red!20}36.86 & \cellcolor{green!20}30.17 & \cellcolor{green!20}\textbf{56.42} & \cellcolor{green!20}\textbf{24.21} & \cellcolor{green!20}39.18 & \cellcolor{green!20}58.95 & \cellcolor{green!20}41.15 & \cellcolor{green!20}30.67 & \cellcolor{green!20}\textbf{25.68} & \cellcolor{green!20}20.46 & \cellcolor{green!20}11.63 \\
\bottomrule
\end{tabular}
}
\caption{Performance on BLEnD dataset. We use \colorbox{green!20}{green} to indicate that our \model exceeds directly prompting the base LLM, and \colorbox{red!20}{red} shading indicates that they do not exceed the base LLM.
We use ISO codes for each country, and the country and code mapping and experiment details can be seen in Appendix \S\ref{subsec: hofstede}.}
\label{BLEnD}
\end{table*}

\subsection{Overall Performance}

\paragraph{\data.} 

In the open-ended question-answering task, as demonstrated in Table~\ref{tab:openculture}, \model surpasses other open-source models and performs comparably to or even better than advanced closed-source models, achieving the highest precision score of 74.73 and cultural F1 score of 72.94 on GlobalCultureQA datasets. 
These results highlight the superior cultural awareness of our proposed \model in addressing open-ended cultural knowledge questions, as well as its ability to effectively generate fine-grained cultural knowledge.

\paragraph{CANDLE500 and CulturalBench.} 

In the cultural knowledge multiple-choice task, as shown in Table~\ref{multi-choice}, \model consistently outperforms base models, achieving improvements of up to 8.8\% on CANDLE500 and 6.6\% on CulturalBench, while also surpassing other open-source models by a large margin. 
However, it still lags behind SOTA models like \texttt{4o}, primarily due to differences in model scale. 
These results demonstrate that our \model effectively enhances the model's cultural capability within the cultural knowledge domain.

\paragraph{BLEnD.} 

When evaluating cultural understanding in local languages, as shown in Table~\ref{BLEnD}, \model outperforms open-source models such as \texttt{Aya} and \texttt{Mistral} in 12 out of 16 countries. 
Notably, the improvements are observed in low-resource language regions, such as Sundanese in West JAVA (increasing from 12.43 to 20.13) and Amharic in Ethiopia (increasing from 8.26 to 12.34). 
Additionally, an intriguing phenomenon emerges: the inherent cultural knowledge distribution within models is highly imbalanced. 
For example, \texttt{Qwen2.5} achieves a score of 60.33 on Chinese, while \texttt{Mistral} scores only 48.31. 
Similarly, \texttt{Aya} attains a score of 66.64 on Indonesian, whereas \texttt{Qwen2.5} scores 49.58. 
This phenomenon likely stems from the differences in pre-training data across various models, highlighting that enhancing a model's cultural competence requires a careful consideration of its internal knowledge architecture and training data composition. In the contrast, \model samples data uniformly across 100+ countries to ensure a balanced representation of diverse culture knowledge with multi-lingual augmentation, which effectively mitigate cultural biases in base models.  
\subsection{Ablation Study}

\begin{table}
    \centering
    \small
    \resizebox{0.8\columnwidth}{!}{
    \begin{tabular}{l|c|c|c}
    \toprule
    \textbf{Model} & \textbf{Precision} & \textbf{Recall} & \textbf{F1} \\
    \midrule
    \model & \textbf{74.73} & \textbf{71.21} & \textbf{72.94} \\
    w/o Critique & 69.54 & 67.63 & 68.57\\
    w/o Multilingual & 70.95 & 70.63 & 70.79\\
    \bottomrule
    \end{tabular}
    }
    \caption{Ablation study on \data.}
    \label{tab:ablation}
\end{table}

We verify the effectiveness of our \model by comparing it with two variant models: \textbf{(1) \model w/o critique}: We remove model-generated answers and corresponding critiques, leaving only golden answers to train our model. 
\textbf{(2) \model w/o multi-lingual}: We exclude the multi-lingual data synthesis stage in our method thus only using mono-English critique data.
As shown in Table~\ref{tab:ablation}, these ablation models both achieve lower scores compared to \model. 
Moreover, removing critique data performs worst in all metrics, which emphasizes the effectiveness of pointing out the weakness in cultural answers, and providing target-aware critique during fine-tuning is crucial for enhancing the cultural ability of models.

In Table~\ref{tab:openculture} and Table~\ref{multi-choice}, we also show the performance of the ablation models: \textit{-DPO} which removes all fine-grained reward data in DPO and \textit{-SFT} which deletes target-aware data in our training dataset.
The performance of these models all decrease on three datasets, with a larger drop in \textit{-SFT}, demonstrating the effectiveness of our training paradigm and the importance of high-quality target-aware SFT data.

\subsection{Analysis of Reward Function}

\begin{figure}[t]
        \centering
	\includegraphics[width=0.5\linewidth]{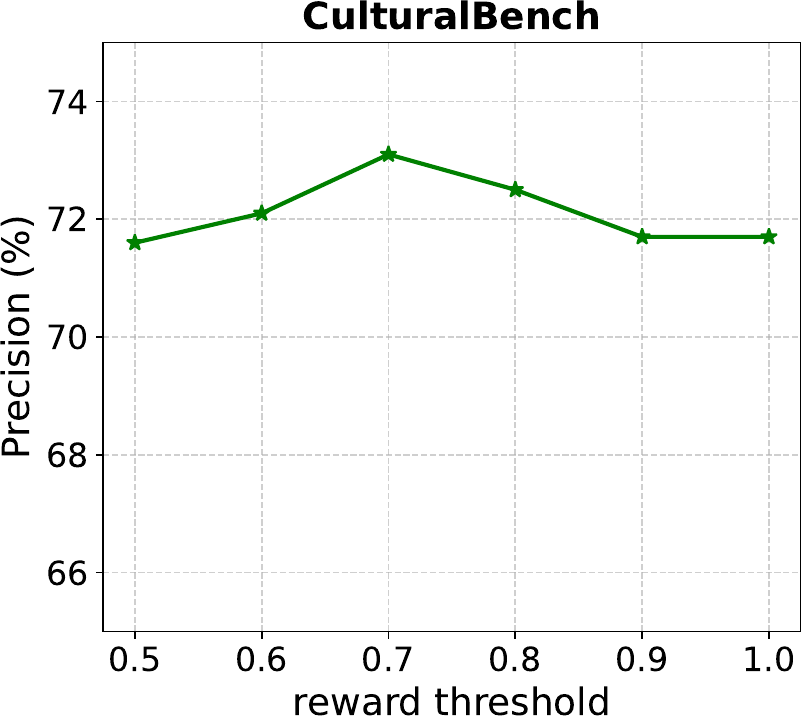}
        \caption{Results of the precision on different reward threshold $S_{f1}$ from 0.5 to 1.0 with an interval of 0.1.}
 \label{fig:reard_threshold}
\end{figure}

We analyze the impact of the reward function by varying the threshold $S_{f1}$ (in Equation \ref{Eq: threshold} $S_{f1}$) for selecting DPO data on the CulturalBench dataset. 
As illustrated in Figure~\ref{fig:reard_threshold}, we observe that setting the threshold to 0.7 yields the best performance for our model, while incorporating higher-performing answers (\eg those rewarded with 0.9) degrades performance. 
A potential explanation for this phenomenon is that DPO benefits from preference pairs with larger differences, as pairs with small differences may hinder the model's ability to identify where errors are likely to occur. 
This finding further validates the effectiveness of our reward function in selecting lower-performing cultural answers, which enhances the model's learning process.

\subsection{Analysis of Cultural Alignment}

We further conduct a cultural alignment evaluation using Hofstede’s cultural dimensions~\cite{hofstede2013vsm}, a well-established framework for quantifying cultural value differences across countries based on data collected from local residents. 
To assess the cultural alignment of LLMs, we prompt them to answer 24 questions from the VSM13 survey~\cite{hofstede2013vsm}, which measures local attitudes toward specific cultural questions. 
We then compute the Euclidean distance across six cultural dimensions between the LLM's responses and human responses. 
% Details descriptions of Hofstede’s cultural dimensions and experimental settings are provided in Appendix \ref{subsec: hofstede}.
Details about Hofstede’s cultural dimensions and experimental settings are provided in Appendix~\ref{subsec: hofstede}.

Figure~\ref{fig:hofstede} presents the results of cultural distances between \texttt{4o}, \texttt{Qwen2.5}, \texttt{Llama3.1}, and our \model, which is based on \texttt{Qwen2.5} and \texttt{Llama3.1}. 
Our findings reveal two key insights:
\textbf{(1)} Our \model outperforms both \texttt{4o} and its base LLM, reducing the cultural distance for \texttt{Llama3.1} from 174.83 to 135.24 and for \texttt{Qwen2.5} from 157.41 to 140.32. 
This demonstrates that \model achieves better cultural value alignment and exhibits superior cultural reasoning capabilities.
\textbf{(2)} Fundamental model abilities, such as math and coding, do not correlate with cultural alignment performance. 
While SOTA LLMs like \texttt{4o} excel in fundamental tasks, they underperform in cultural value alignment compared to smaller models. 
This discrepancy may stem from the unbalanced cultural knowledge in the training data of SOTA LLM, which can skew their value systems.
These results highlight the importance of reducing cultural bias and developing models that ensure equitable cultural representation.

\begin{figure}[t]
        \centering
	\includegraphics[width=0.8\linewidth]{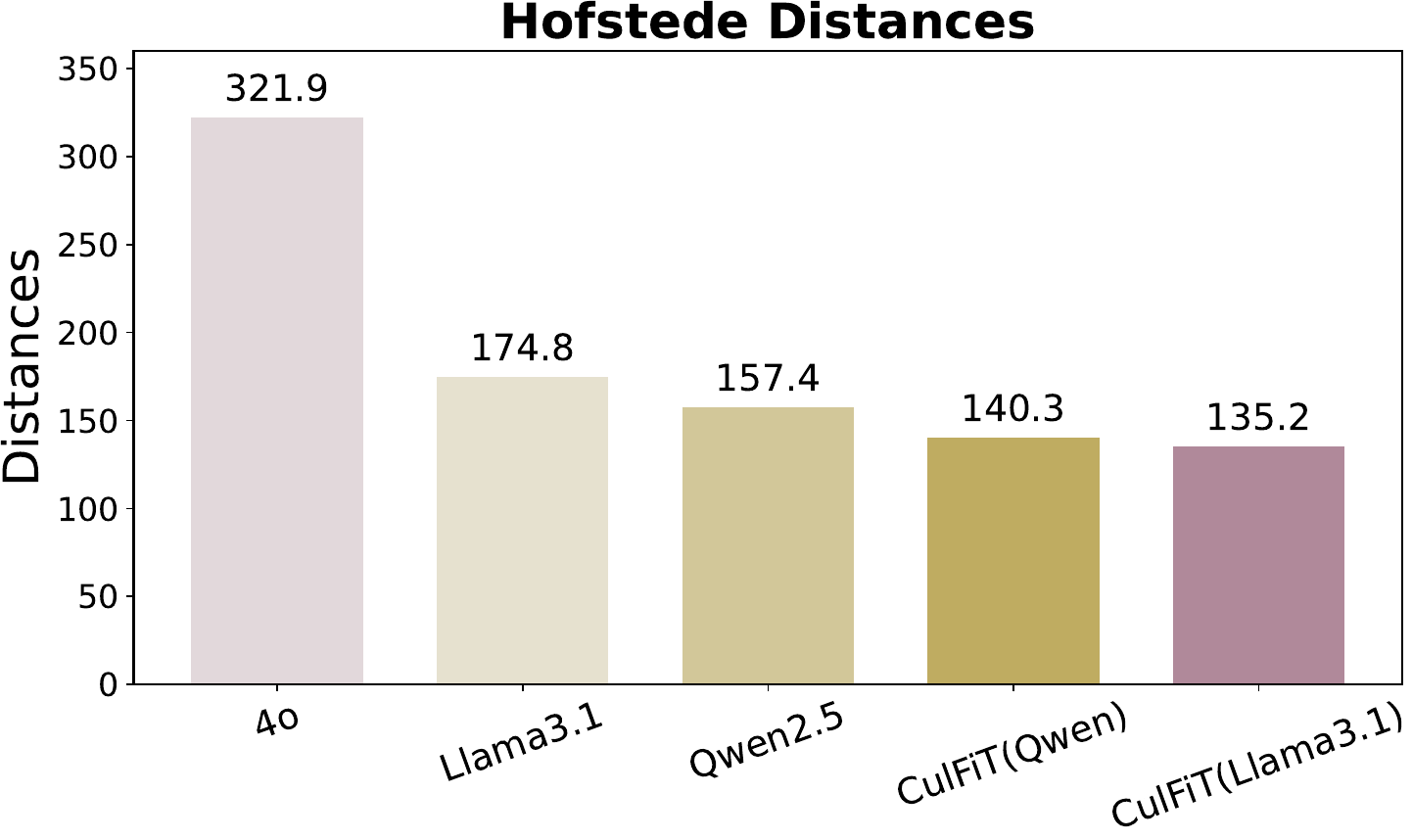}
        \caption{Comparison in terms of Hofstede distance.}
 \label{fig:hofstede}
\end{figure}

\subsection{Analysis of Multilingual Data}

\begin{table}
\centering
\small
\begin{tabular}{l|c|c}
\toprule
\textbf{Model} & \textbf{MMLU} & \textbf{MMMLU} \\
\midrule
\texttt{Llama3.1} & 49.6 & 52.0\\
\texttt{Qwen2.5} & 66.2 & 63.9 \\
\model (Llama3.1) & \textbf{50.1} & \textbf{62.5} \\
\model (Qwen2.5) & \textbf{66.4} & \textbf{67.9} \\
\bottomrule
\end{tabular}
\caption{Precision scores on multi-lingual scenario.}
\label{robustness}
\end{table}

\begin{figure}[t]
        \centering
	\includegraphics[width=0.5\linewidth]{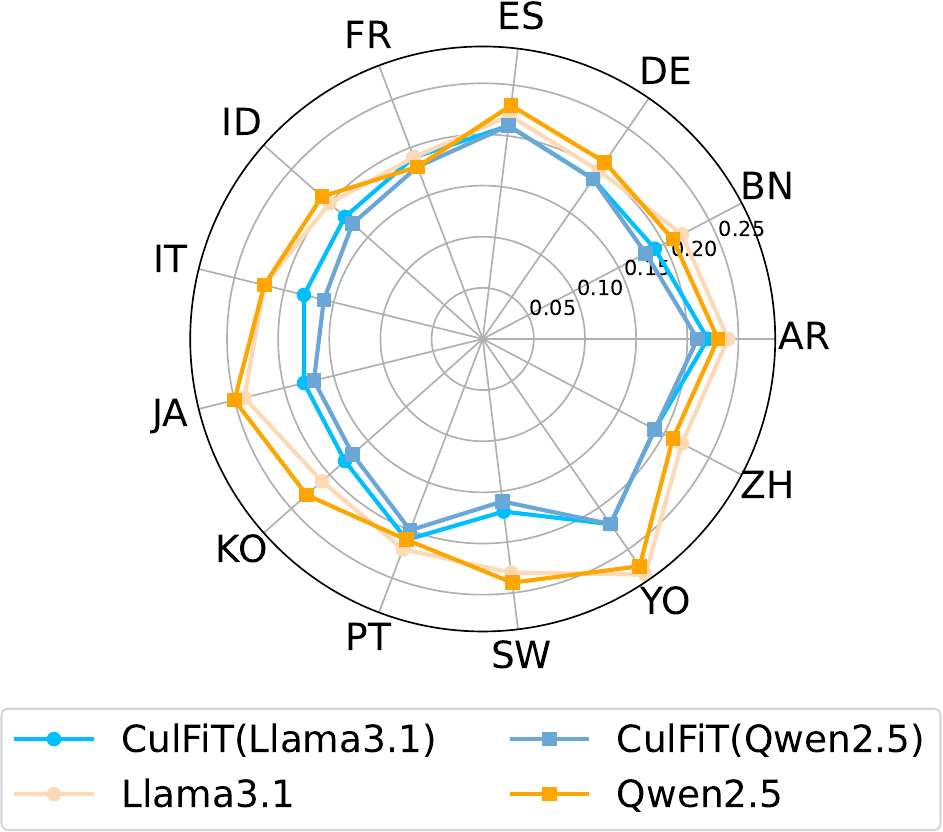}
        \caption{Results on multi-lingual inconsistent rate.}
 \label{fig:robusness}
\end{figure}

To investigate the effectiveness of our \model in multilingual settings, we conduct experiments on the MMLU~\cite{hendrycks2020measuring} and MMMLU~\cite{wang2024mmlu} dataset which translates MMLU into 14 languages. 
We randomly select 150 English questions related to cultural domains such as world religions, human sexuality, and sociology. 
Table~\ref{robustness} reports the precision of our \model and its base LLM \texttt{Llama3.1}, where our model outperforms \texttt{Llama3.1} by a large margin (10.5\% in MMMLU) when answering the same cultural questions while maintaining comparable or even superior performance in MMLU.

To validate the robustness of our \model in the multilingual scenario, we count the inconsistent responses between two models on two datasets (\aka MMLU and MMMLU) and group the results by language. 
Figure~\ref{fig:robusness} illustrates that our model exhibits a lower inconsistent error rate than base models across all 14 languages, demonstrating excellent robustness.
\begin{table}
\centering
\resizebox{1\columnwidth}{!}{
\begin{tabular}{l|c|c|c}
\toprule
\textbf{Model} & \textbf{CSQA} & \textbf{Hellaswag} & \textbf{MMLU-pro} \\
\midrule
\texttt{Llama3.1} & 70.1 & 71.5 & 36.8 \\
\model (Llama3.1) & 73.1(+3.0) & 72.7(+1.2) & 38.7(+1.9)\\
\midrule
\texttt{Qwen2.5} & 80.3 & 75.9 & 47.0 \\
\model (Qwen2.5) & \textbf{80.6(+0.3)} & \textbf{77.5(+1.6)} & \textbf{47.1(+0.1)} \\
\bottomrule
\end{tabular}
}
\caption{Comparison of general abilities of between \model and base LLM.}
\label{general ability}
\end{table}

\subsection{Discussion of General Capability}

To evaluate the generalization ability of our method and mitigate the risk of catastrophic forgetting, we conduct experiments on commonsense and reasoning datasets, including CSQA~\cite{talmor2018commonsenseqa}, Hellaswag~\cite{zellers2019hellaswag}, and MMLU-pro~\cite{wang2024mmlu}. 
As shown in Table~\ref{general ability}, our models consistently outperform the original LLM across all three tasks, demonstrating that integrating culture-related knowledge by using our proposed \model not only enhances culture-related knowledge but also improves general reasoning capabilities and prevents catastrophic forgetting.

\section{Conclusion}

In this paper, we introduced \fullmodel (\model), a novel cultural-aware training paradigm that addresses cultural bias in large language models (LLMs) through target-aware multilingual data synthesis and fine-grained reward modeling. 
Our approach enhances cultural sensitivity and robustness across diverse linguistic and cultural contexts.
Experiments on our newly proposed \data benchmark and three cultural knowledge benchmarks show that \model outperforms existing open-source models and competes with state-of-the-art closed-source models. 
% Ablation studies confirm the importance of target-aware critique and multilingual data synthesis in improving cultural alignment.
Analysis using Hofstede’s cultural dimensions reveals that \model achieves better cultural value alignment than base models and advanced LLMs like GPT-4o. 
% Multilingual robustness experiments further demonstrate consistent performance across 14 languages.
% Our work advances cultural-aware LLM training and provides a framework for evaluating cultural alignment, paving the way for more inclusive and culturally sensitive AI technologies.

% \clearpage

\section*{Limitations}

While \model shows significant improvements, it still faces challenges in fully capturing the fine-grained cultural knowledge of low-resource languages due to limited training data.
Another minor limitation is the computational cost associated with generating and processing multilingual critique data, which could be a bottleneck for smaller research teams.

\section*{Ethical Considerations}

Despite ongoing efforts to reduce cultural bias, large language models (LLMs) can still unintentionally reinforce stereotypes or present inaccurate portrayals of certain cultures. This often stems from biases embedded in the data they are trained on, which may reflect dominant cultural narratives or historical inequalities. As a result, the outputs generated by these models may marginalize underrepresented voices or misrepresent diverse communities. Addressing these issues is a critical ethical responsibility to ensure fairness, inclusivity, and respectful representation in AI systems.

\section*{Acknowlegements}
This work was supported by the National Key R\&D Program of China 2024YFE0111800, the National Natural Science Foundation of China (62032001, 62432002, 62406061, and T2293773), and the Natural Science Foundation of Shandong Province (ZR2023QF159).

%\bibliography{anthology,custom}
\bibliography{custom}
\clearpage
\section{Appendix}
\subsection{Training and Implementation Details}
\label{subsec: training detail}
In the supervised fine-tuning(SFT) stage, we use CANDLE and CultureAtlas as seed data to train LLMs and in direct preference optimization(DPO) stage we adopt CultureBank as source data. We arrange the data in a dataflow of Question->Original Answer->Critique->Golden answer, with Golden answer as output and others as output.
We finally get 25344 QA pairs with critique in English and 20,140 pairs in other languages, spanning 3026 topics and 24 languages in supervised-finetuning stage and 16334 preference pairs in Direct preference Optimization stage.

We train our models on 8 NVIDIA L40s and train the model for 1000 steps of batch size 16 on every stage. We select a learning rate of 1e-5 in SFT stage and 5e-6 in DPO stage with a warmup ratio of 0.1. For parameter efficiency, all training process use LoRA with a rank of 16.

% In supervised fine-tuning stage, we only use CANDLE and CultureAtlas as seed data and go through our method. 
% We finally get 25344 QA pairs with critique in English and 20,140 pairs in other languages, spanning 3026 topics and 72 languages. 
% We then rearrange the collected data into sequences of sentences as follows: Instruction → Original answer → Critique → Golden answer to train the model.
% In Direct Prefence Optimization stage, we further increase the cultural ability of our model using CultureBank as source data. 
% Notably, we integrated SFT loss of DPO positive examples during training as an approximate substitute for a regularization term. 
% Despite this, we also design \textbf{Cultural F1 Score} as reward function to fully assess the model's cultural QA ability which will be discussed in \ref{sec: Evaluation Metric} 

% \subsection{Implementation Detail and Baselines} 

We use greedy decoding for multi-choice questions and tak temperature of 0.7 for other tasks. We use one-shot prompting on our GlobalCultureQA dataset during inference and adopt zero-shot prompting strategy for other tasks.

\subsection{Details on GlobalCulture QA}
\label{subsec: GlobalCultureQA}
\begin{figure}[h]
        \centering
	\includegraphics[width=1\linewidth]{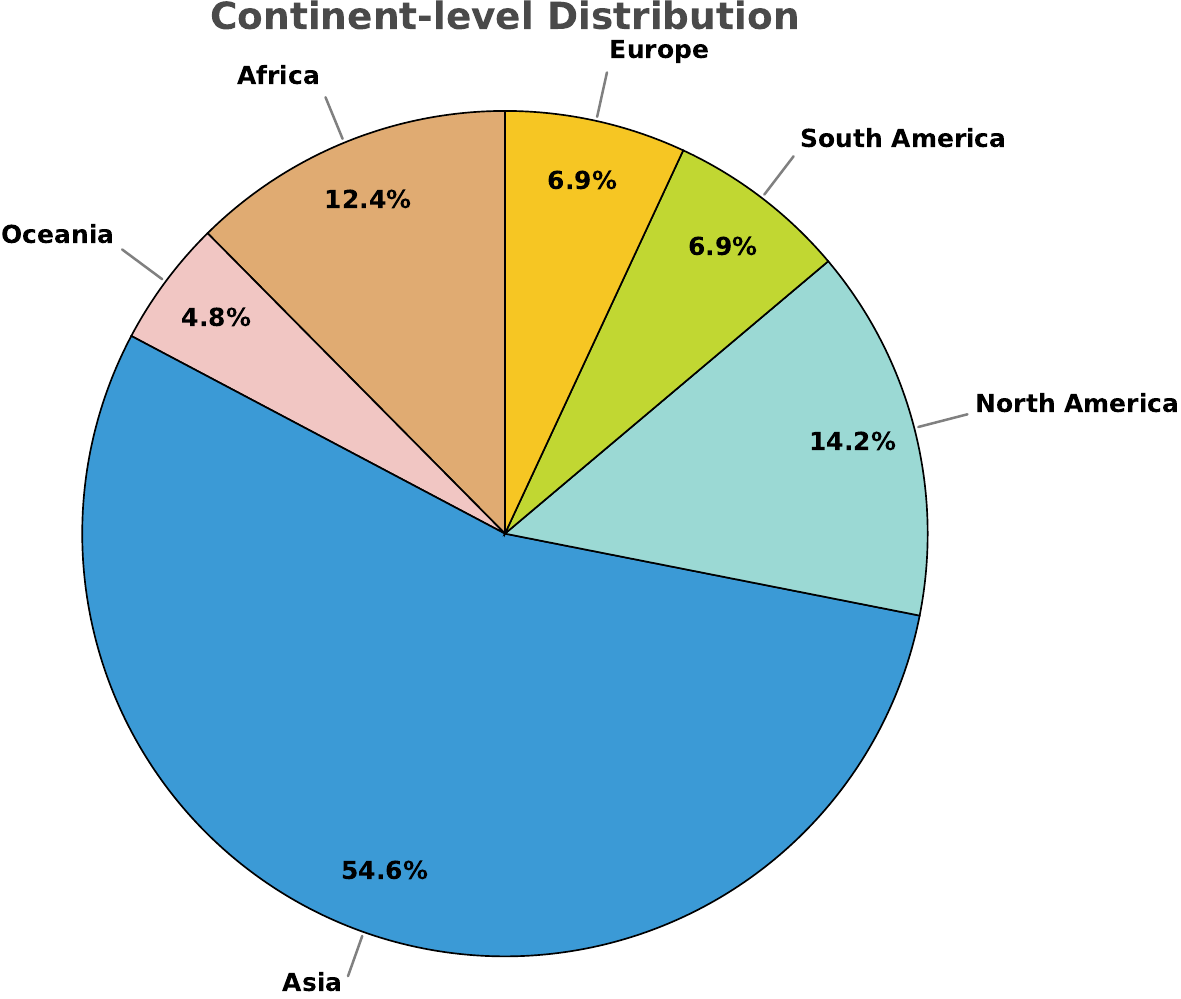}
        \caption{The continent level distribution on our GlobalCultureQA}
 \label{fig:pie_benchmark}
\end{figure}

\begin{figure}[h]
        \centering
	\includegraphics[width=1\linewidth]{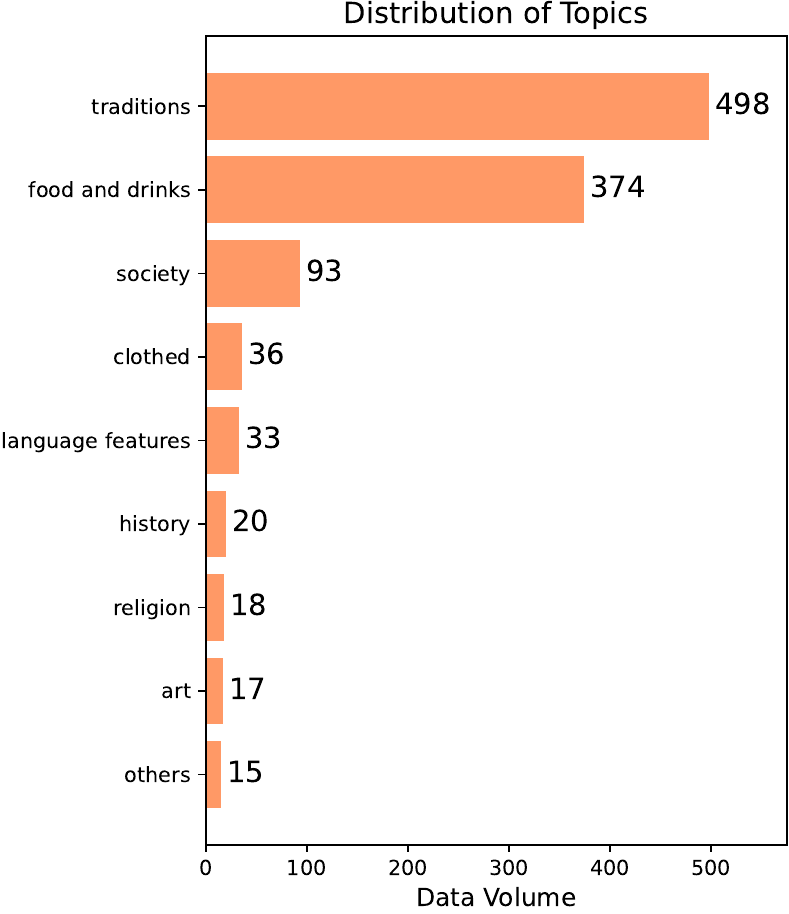}
        \caption{Distribution of overall topics. We classify the specific topics into 9 categories and report the volume of each category}
 \label{fig:bar_benchmark}
\end{figure}

We provide an example of our benchmark's question:
\begin{tcolorbox}[colback=black!1!white, colframe=black!57!white, boxsep=1pt, left=1pt, right=1pt, top=1pt, bottom=1pt, breakable]

How do the enforcement and interpretation of hijab regulations differ between regions such as Gaza, Iran, Indonesia, and Saudi Arabia, and what social tensions do these differences reflect?
\end{tcolorbox}

and we provide an example of the corresponding golden answer:
\begin{tcolorbox}[colback=black!1!white, colframe=black!57!white, boxsep=1pt, left=1pt, right=1pt, top=1pt, bottom=1pt, breakable]

The enforcement and interpretation of hijab regulations vary significantly across regions such as Gaza, Iran, Indonesia, and Saudi Arabia, reflecting complex social tensions. In Gaza, members of the Palestinian jihadist group Unified Leadership have rejected policies mandating the hijab, even targeting those who attempt to enforce such rules. In Iran, the shift from a ban on veils in 1936 to mandatory Islamic dress following the 1979 Islamic Revolution established a stringent legal framework, leading to severe penalties for women not adhering to hijab requirements. This has resulted in fluctuating enforcement practices, with women often challenging these restrictions. In Indonesia, particularly in the province of Aceh, local Sharia law mandates that Muslim women wear hijab in public, showcasing a more localized interpretation of dress codes. In Saudi Arabia, while the government requires women to cover their hair and wear full-body garments, enforcement has been inconsistent, leading to criticism of the religious police for their actions, notably hindering the rescue of schoolgirls in 2002 due to their attire. Overall, these differing practices highlight ongoing tensions and varied interpretations of religious and cultural norms across societies, reflecting broader social dynamics and conflicts over gender, identity, and religious expression.
\end{tcolorbox}

\subsection{Details On BLEnD}
\label{subsec: BLEnD}

When conducting experiments on BLEnD, we adpot a system prompt \texttt{'You are a helpful \{country\} AI chatbot that know the culture of \{country\} very well. You task is to answer the question about \{country\} in \{language\}.'}

We also choose the results of 'pers-3' prompt described in the original paper: \texttt{'You are a person from \{country\} who is trying to explain your country's culture to a foreigner. Answer the following question, providing a single answer without any explanations.'}

Table \ref{tab:country_lang} shows the mapping of country and ISO code, with the corresponding answers of each country.
\begin{table}[ht!]
\centering
\begin{tabular}{ll|ll}
\toprule
\textbf{Country/Region} & \textbf{Code} & \textbf{Language}  \\ \midrule
United States   & US & \multirow{2}{*}{English}  \\ 
United Kingdom  & GB &                            \\ \midrule
China           & CN & Chinese                    \\ \midrule
Spain           & ES & \multirow{2}{*}{Spanish}  \\ 
Mexico          & MX &                         \\ \midrule
Indonesia       & ID & Indonesian                \\ \midrule
South Korea     & KR & \multirow{2}{*}{Korean}   \\ 
North Korea     & KP &                           \\ \midrule
Greece          & GR & Greek                    \\ \midrule
Iran            & IR & Persian                    \\ \midrule
Algeria         & DZ & Arabic                   \\ \midrule
Azerbaijan      & AZ & Azerbaijani               \\ \midrule
West Java       & JB & Sundanese                \\ \midrule
Assam           & AS & Assamese                 \\ \midrule
Northern Nigeria & NG & Hausa                    \\ \midrule
Ethiopia        & ET & Amharic                  \\ 
\bottomrule
\end{tabular}
\caption{The details of country and ISO code mapping with their corresponding languages}
\label{tab:country_lang}
\end{table}

\subsection{Details On Hofstede Cultural Dimentions}
\label{subsec: hofstede}
This survey identified six dimensions of national culture: Power Distance Index (PDI), Individualism vs. Collectivism (IDV), Masculinity vs. Femininity (MAS), Uncertainty Avoidance Index (UAI), Long-Term Orientation vs. Short-Term Orientation (LTO) and Indulgence vs. Restraint (IND). VSM 2013 is an authoritative and famous cultural questionnaire devised by Hofstede and is widely used. In this experiment, we evaluate the cultural alignment of our models on $9$ cultures(Arabic, Bangladesh, Chinese Germany, Korean, Portuguese, Brazil, Argetina and Turkish) and calculate the average distances of all countries.
% country_dict = {'Arabic': 'Jordan', 'Bengali': 'Bangladesh', 'Chinese': 'China', 'Germany': 'Germany',
%                 'Korean': 'Korea South', 'Portuguese': 'Brazil', 'Spanish': 'Argentina', 'Turkish': 'Turkey'}
To be specific, the VSM 2013 have $24$ questions in total. The computation of six cultural dimensions is based on the following formulas:
\begin{equation}
PDI=35(\mu_{Q7}-\mu_{Q2})+25(\mu_{Q20}-\mu_{Q23}) + C_{PDI}
\end{equation}
\begin{equation}
IDV=35(\mu_{Q4}-\mu_{Q1})+35(\mu_{Q9}-\mu_{Q6}) + C_{IDV}
\end{equation}
\begin{equation}
MAS=35(\mu_{Q5}-\mu_{Q3})+25(\mu_{Q8}-\mu_{Q10}) + C_{MAS}
\end{equation}
\begin{equation}
UAI=40(\mu_{Q18}-\mu_{Q15})+25(\mu_{Q21}-\mu_{Q24}) + C_{UAI}
\end{equation}
\begin{equation}
LTO=40(\mu_{Q13}-\mu_{Q14})+25(\mu_{Q19}-\mu_{Q22}) + C_{LTO}
\end{equation}
\begin{equation}
IVR=35(\mu_{Q12}-\mu_{Q11})+40(\mu_{Q17}-\mu_{Q16}) + C_{IVR}
\end{equation}

\noindent where $\mu$ means the average of all the answers to each question. $C$ is constants that can be used to adjust to scores to fit a range between $0$ and $100$ or anchor new data to Hofstede's old dataset~\citep{hofstede2013vsm}. During experiment, we convert the questions into the multi-choice format, with a \texttt{'You are a \{culture\} chatbot that know \{culture\} very well. Now your task is to represent the people in {culture} and answer the following question. Please be sure that you should only consider the culture of \{culture\} when answering the question.'} system prompt.

We get the Euclidean distance of the gaps from six cultural dimensions as the metric $Distance$. Larger $Distance$ means weak cultural understanding ability of LLMs, and vice versa. 
\begin{equation}
\label{eq-distance}
Distance = \sqrt{\sum (d_{model} - d_{hofstede})^2}, 
% \forall d \in \{PDI,IDV,MAS,UAI,LTO,IVR\}.
\end{equation}

\begin{figure*}[h]
    \centering
    \includegraphics[width=1\linewidth]{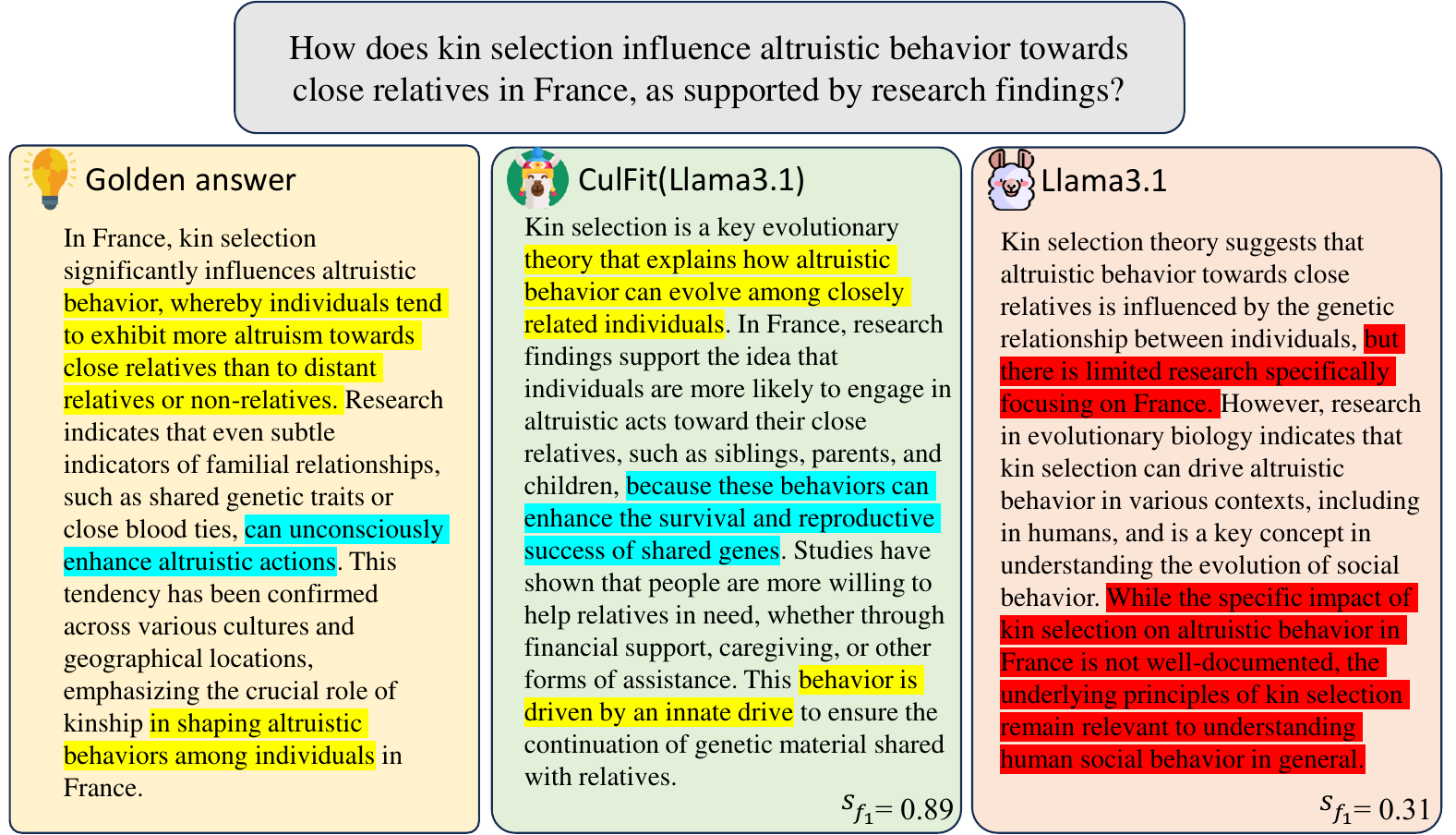}
    \caption{A case study on the results of \data. We use \colorbox{yellow}{yellow} to indicate the the part that corresponds with golden answer, \colorbox{blue!20}{blue} to show the extensive content compared to golden answer and \colorbox{red}{red} to highlight vague and uncultural answers.}
    \label{fig:case_study}
\end{figure*}
\subsection{Case Study}
\label{subsec: Case Study}
As shown in Figure \ref{fig:case_study}, we compare the answer of our proposed \model and \texttt{Llama3.1}. We use \colorbox{yellow}{yellow} to indicate the part that corresponds with golden answer, and We use \colorbox{blue!20}{blue} to highlight content that is more extensive compared to the golden answer. \colorbox{red}{Red} highlights indicate responses that are vague compared to the golden answer and fail to provide a corresponding answer. In our \model's answer, we have parts that precisely reflect the golden answer(\texttt{\color{yellow}theory that explains how altruistic behavior can evolve among closely related individuals}) and have contents that extend the cultural knowledge to a more nuanced extent(\texttt{\color{blue}because these behaviors can enhance the survival and reproductive success of shared genes}). On the contrast, the original \texttt{Llama3.1} just gives vague and incorrect answers like \texttt{\color{red}but there is limited research specifically focusing on France.}, which hinders the cultural nuances in these sentences. We attribute this to the target-aware data training because we force model to capture the 'target' in the question and thus avoid generating bad answers like \texttt{\color{red}While the specific impact of kin selection on altruistic behavior in France is not well-documented}. Additionally, our \model get a cultural f1 score $S_{f1}$ of 0.89, while \texttt{Llama3.1} only obtain 0.31, which is correlated with the analysis above, demonstrating the stability and fairness of our evaluation metric.

\subsection{Human Evaluation}
We performed a model comparison study involving 100 randomly sampled questions. Four native speakers (1 PhD and 3 Master's students proficient in both Chinese and English) participated in this single-blind evaluation. The annotators achieved an impressive inter-annotator agreement (Kappa score = 0.86), demonstrating high consistency in assessing cultural awareness, readability, and question relevance. Results in Table\ref{tab:human_study} show our CulFit (Llama3.1) model significantly outperforms the base model, aligning better with human preferences across all evaluation dimensions. 
\begin{table}
\centering
\small
\begin{tabular}{l|c|c|c}
\toprule
\textbf{Model} & \textbf{Win} & \textbf{Tie} & \textbf{Lose} \\
\midrule
\model(Llama3.1) vs \texttt{Llama3.1} & 38 & 46 & 16\\
\model (Llama3.1) vs \texttt{gpt-4o} & 25 & 50 & 24 \\
\bottomrule
\end{tabular}
\caption{Human study of our model vs Llama3.1 and gpt-4o}
\label{tab:human_study}
\end{table}

\subsection{Prompts For Data Synthesis}
The prompt for cultural question generation based on cultural knowledge:
\begin{tcolorbox}[colback=black!1!white, colframe=black!57!white, boxsep=1pt, left=1pt, right=1pt, top=1pt, bottom=1pt, breakable]

You are a helpful expert in generating cultural-aware quetions through cultural knowledge. You are privided with a piece of cultural knowledge and the background of the cultural knowledge. Your task is to generate a single question based on the cultural knowledge that is given to you. The input form is encoded as JSON format, and below is its JSON fields:
\{"cultural\_group": "", "topic": "", "source": "", "cultural\_knowledge": ""\} 

the detailed explanation of the fields are as follows:\\
-cultural\_group: the country or the cultural group where the cultural knowledge is from\\
-topic: the topic of the cultural knowledge\\
-source: the source of the cultural knowledge\\
-cultural\_knowledge: the cultural knowledge that is provided to you, which should pay most attention\\

Please strictly follow the following rules:
1. Factuality: Your question should only stems from the cultural knowledge that is provided to you and you shouldn't add other knowledge to your generated question.
2. Specificity: Your question should cover the main idea of the cultural knowledge and should be comprehensive, but not too broad. Try to specific the question with the cultural knowledge and do not ask too general questions.
3. Coverage: You should carefully understand the cultural knowledge and extract the cultural knowledge points as much as possible. And use these cultural knowledge ponints to formulate your question. 
\end{tcolorbox}

\label{subsec: prompts}
The prompt for answer generation process:
\begin{tcolorbox}[colback=black!1!white, colframe=black!57!white, boxsep=1pt, left=1pt, right=1pt, top=1pt, bottom=1pt, breakable]

You are a helpful consultant for a cultural knowledge question answering scenario. You are given the following question and its cultural knowledge. Your task is to generate a culturally-aware answer to the question based on the cultural knowledge.

Remember, your answer should be encoded in JSON format. The detailed explanation of the fields is as follows:

\texttt{\{"answer": "", "cultural\_group": "", "language": "", "topic": ""\}}

\textbf{answer}: your answer to the question \\
\textbf{cultural\_group}: the country or the cultural group your answer points to \\
\textbf{language}: the language that the cultural group mainly speaks \\
\textbf{topic}: the main topic of your answer

--------

Notably, the question stems from the cultural knowledge, so your answer should also be based on the provided cultural knowledge. You should always follow the instructions and directly answer the questions that are provided to you.\\
<example\_start> \\
... \\
<example\_end> \\
Remember, your answer should correlate with the cultural knowledge . You should only return the answer.

Your Answer:

% \caption{the prompt for answer generation process}
\end{tcolorbox}

The prompt for target-aware critique generation:
\begin{tcolorbox}[colback=black!1!white, colframe=black!57!white, boxsep=1pt, left=1pt, right=1pt, top=1pt, bottom=1pt, breakable]

You are an expert reviewer for a cultural knowledge question answering system. You have plenty of cultural knowledge in \{\}.

You are given a JSON object and the detailed explanation of the fields are as follows: \\
\{"question":"","grounded\_answer":"", "answer\_to\_critique":"", "grounded\_answer\_knowledge\_points":"", “knowledge\_points\_to\_critique”: ""\}\\
-question: the cultural question that is given to you\\
-grounded\_answer: the grounded answer to the question, which is the reference answer\\
-answer\_to\_critique: the answer that you should critique\\
-grounded\_answer\_knowledge\_points: the knowledge points extracted from the grounded answer, each knowledge point is a single sentence and is seperated with a comma in a list\\
-knowledge\_points\_to\_critique: the knowledge points extracted from the answer\_to\_critique, each knowledge point is a single sentence and is seperated with a comma in a list

You should compare the grounded\_answer\_knowledge\_points  and the answer\_knowledge\_points and provide a detailed critique based on the comparison. And your critique should based on the principles below:
1. Correctness: Be sure to point out any factual inaccuracies or errors in the answer\_to\_critique and provide corrections based on the grounded\_answer\_knowledge\_points.
2. Comprehensiveness: The answer\_to\_critique should cover the main points of the grounded\_answer and should not miss any key information, if the answer\_to\_critique miss the cultural knowledge points, you should say "not addressed clearly" between the comparison.
3. Stability: If the grounded\_answer\_knowledge\_points and the knowledge\_points\_to\_critique are mainly the same, you should say "Roughly the same" in your critique.
4. Point by point: You should compare the grounded\_answer\_knowledge\_points and knowledge\_points\_to\_critique point by point and provide your critique based on the comparison. Between the comparison, you should choose the most relevant knowledge\_points\_to\_critique from the list while comparing the grounded\_answer\_knowledge\_points.

You should always follow the instructions and carefully compare the grounded\_answer\_knowledge\_points and the answer\_knowledge\_points point by point and provide your critique.
Remember, you should directly compare the grounded\_knowledge\_points and knowledge\_points\_to\_critique and point out the flaws made by knowledge\_points\_to\_critique.

And your critique should be encoded in JSON format, with each unit be a small JSON format , which will be like:\\
\{\\
    \{\\
        "grounded\_answer\_knowledge\_points": "",\\
        "knowledge\_points\_to\_critique":""\\
        "Critique": ""\\
    \},\\
    \{\\
        "grounded\_answer\_knowledge\_points": "",\\
        "knowledge\_points\_to\_critique": ""\\
        "Critique": ""\\
    \},\\
    .....\\
\}\\
You answer:
\end{tcolorbox}

\subsection{Examples Of Critique And Answer}
We also elaborate an example of meta critique:

\begin{tcolorbox}[colback=black!1!white, colframe=black!57!white, boxsep=1pt, left=1pt, right=1pt, top=1pt, bottom=1pt, breakable]

    \{\\
        "grounded\_answer\_knowledge\_points": "In Chinese culture, eggs symbolize fertility and new life.", \\
        "knowledge\_points\_to\_critique": "In Chinese culture, eggs symbolize fertility and new life due to their ability to hatch.", \\
        "Critique": "Roughly the same"  \\
    \},\\
    \{\\
        "grounded\_answer\_knowledge\_points": "Eggs represent blessings and auspiciousness in Chinese culture.",\\
        "knowledge\_points\_to\_critique": "Not addressed clearly.",\\
        "Critique": "The answer\_to\_critique does not mention the aspect of blessings and auspiciousness associated with eggs, which is an important cultural significance. Including this detail would enhance the comprehensiveness of the response."\\
    \},\\
    \{\\
        "grounded\_answer\_knowledge\_points": "Eggs play an integral role in many aspects of Chinese life.",\\
        "knowledge\_points\_to\_critique": "Contradictory.",\\
        "Critique": "The answer\_to\_critique does not encapsulate the integral role that eggs play in various aspects of Chinese life, which is a key point. Acknowledging this would provide a holistic view of the cultural significance of eggs."\\
    \}\\

\end{tcolorbox}

We also elaborate an example of the Critique $C$: \\
\begin{tcolorbox}[colback=black!1!white, colframe=black!57!white, boxsep=1pt, left=1pt, right=1pt, top=1pt, bottom=1pt, breakable]

The critique identifies several key aspects of Chinese cultural significance related to eggs that are either missing or inadequately addressed in the answers. While some points align closely, many important cultural dimensions are overlooked. Specifically, the aspects of blessings and auspiciousness associated with eggs, the gifting of eggs during Chinese New Year, and their involvement in rituals and culinary practices are not clearly articulated. Additionally, the use of eggs in fortune-telling, their culinary significance in traditional dishes like mooncakes, and the symbolic representation of the yolk are omitted. Unique customs, such as boiling eggs in urine and using eggshells for folk art, as well as the broader role of eggs in Chinese life, also lack sufficient discussion. Including these details would enhance the depth and richness of the response, providing a more comprehensive understanding of the cultural significance of eggs in Chinese traditions.
\end{tcolorbox}

Additionally, here is an example of our question:
\begin{tcolorbox}[colback=black!1!white, colframe=black!57!white, boxsep=1pt, left=1pt, right=1pt, top=1pt, bottom=1pt]

How does the integration of high-quality materials, such as English wool blends, and traditional Chinese techniques in the production of hand-tufted rugs like Whisper reflect the rich textile heritage of China?
\end{tcolorbox}

And here is an example of the corresponding answer:
\begin{tcolorbox}[colback=black!1!white, colframe=black!57!white, boxsep=1pt, left=1pt, right=1pt, top=1pt, bottom=1pt, breakable]

The integration of high-quality materials, such as English wool blends, and traditional Chinese techniques in the production of hand-tufted rugs like Whisper reflects the rich textile heritage of China by showcasing the country's exceptional craftsmanship and adaptability. The use of fine English wool enhances the quality of the rugs, while traditional Chinese techniques ensure that the artistry and cultural significance of the textile production are preserved. This blending of premium materials and age-old methods illustrates how China not only honors its textile heritage but also embraces global influences to create products that stand out in the market, reflecting both tradition and innovation."
\end{tcolorbox}

We also exhibt the example of answer unit:
\begin{tcolorbox}[colback=black!1!white, colframe=black!57!white, boxsep=1pt, left=1pt, right=1pt, top=1pt, bottom=1pt, breakable]

\{\\
"knowledge\_points":\\
"The production of hand-tufted rugs like Whisper integrates high-quality materials such as English wool blends with traditional Chinese techniques.",\\
"This integration reflects the rich textile heritage of China.",\\
"The combination of modern materials and ancient techniques showcases the mastery of Chinese artisans.",\\
"The use of English wool blends contributes softness, durability, and stain resistance to the rugs.",\\
"Traditional Chinese techniques like hand-tufting and natural dyeing maintain the rugs' cultural and aesthetic value.",\\
"The fusion of old and new in rug production demonstrates China's long history of textile innovation.",\\
"Chinese textile production adapts to changing times while remaining true to cultural roots."\\\}
\end{tcolorbox}

\subsection{Evaluation Example}

We first display the prompt for our fine-grained evaluation process:
\begin{tcolorbox}[colback=black!1!white, colframe=black!57!white, boxsep=1pt, left=1pt, right=1pt, top=1pt, bottom=1pt, breakable]
You are an expert evaluator for a cultural knowledge question answering system. You are given a piece of cultural knowledge point and a list of reference cultural knowledge. Your task is to evaluate whether the given cultural knowledge point satisfies one of the reference cultural knowledge points and give a concise explanation.\\
Here are some examples and explanations: \\
</example> \\
<example/> \\

Remember, Your output should first generate 'Yes' or 'No', and give a concise explanation of your evaluation. \\
If your answer is "Yes", your explanation should specifically incorporate the given cultural knowledge point satisfies which reference cultural knowledge point.\\
cultural knowledge points:\\
\{\}\\
reference cultural knowledge points:\\
\{\}\\
Your output:
\end{tcolorbox}

We then display the 'Yes' case of the evaluation process with explanation:
\begin{tcolorbox}[colback=black!1!white, colframe=black!57!white, boxsep=1pt, left=1pt, right=1pt, top=1pt, bottom=1pt, breakable]
cultural knowledge points:\\
"The centers aim to improve literacy rates among Afghan citizens, particularly women and children."\\

reference cultural knowledge points:\\
"Lincoln learning centers in Afghanistan improve literacy rates among Afghan citizens.", \\
"These centers were established in response to low literacy rates in Afghanistan.", \\
"Lincoln learning centers serve as educational hubs providing English language classes, library facilities, Internet connectivity, and counseling services.",\\ 
"The initiative aims to reach at least 4,000 Afghan citizens each month at each location.", \\
"Literacy courses are mandatory for the military and national police forces in Afghanistan.",\\
 "The initiative reflects a broader commitment to enhancing literacy levels across Afghanistan.", \\
 "Educational programs at the centers promote an understanding of American culture.", \\
 "The primary languages spoken in the Lincoln learning centers are Dari and Pashto."\\

Your output:\\
Yes\\

explanation: the cultural knowledge point "The centers aim to improve literacy rates among Afghan citizens, particularly women and children." is similar to the reference cultural knowledge point "These centers were established in response to low literacy rates in Afghanistan.", so the output is Yes.

\end{tcolorbox}

And a 'No' case for the evaluation with explanation:
\begin{tcolorbox}[colback=black!1!white, colframe=black!57!white, boxsep=1pt, left=1pt, right=1pt, top=1pt, bottom=1pt, breakable]
cultural knowledge points:\\
"Basketball is gaining popularity in Afghanistan and is enjoyed by both men and women."\\

reference cultural knowledge points:\\
"The Afghan Sports Federation was established in 1922.",\\
"The Afghan Sports Federation promotes sports like football and basketball in Afghanistan.",\\
"The federation is responsible for developing, organizing, and overseeing various sports in Afghanistan.",\\
"Afghanistan's national football team qualified for the 2014 FIFA World Cup.",\\
"The qualification for the 2014 FIFA World Cup was a significant milestone for Afghan football.",\\
"The Afghan Sports Federation faces challenges such as financial constraints and infrastructure limitations.",\\
"Ongoing conflicts in Afghanistan have impacted the development of sports.",\\
"The history of the national football team reflects Afghanistan's turbulent past.",\\
"Many players and coaches of the national football team have fled Afghanistan due to conflict or persecution.",\\
"The Taliban banned sports during their rule from 1996 to 2001.",\\
"The ban on sports during Taliban rule hindered the progress of the national football team.",\\
"The national football team has shown resilience despite numerous challenges.",\\
"Players like Zohib Islam Amiri and Faisal Hamidi have represented Afghanistan in international competitions.",
"The 2021 Taliban takeover has raised concerns about the future of sports in Afghanistan."\\

Your output:\\
No\\

explanation: the cultural knowledge point "Basketball is gaining popularity in Afghanistan and is enjoyed by both men and women" is not addressed clear in the reference cultural knowledge points, so the output is No.
\end{tcolorbox}
% \appendix

% \section{Example Appendix}
% \label{sec:appendix}

% This is an appendix.

\end{document}